# EnTRPO: Trust Region Policy Optimization Method with Entropy Regularization

**Sahar Roostaie[a], Mohammad Mehdi Ebadzadeh[b,*]**

[a] PhD student at Amirkabir University of Technology, Hafez street, Tehran, Iran
[b] Professor at department of Computer Engineering, Amirkabir University of Technology, Hafez street, Tehran, Iran

## ABSTRACT

Trust Region Policy Optimization (TRPO) is a popular and empirically successful policy search algorithm in reinforcement learning (RL). It iteratively solved the surrogate problem which restricts consecutive policies to be 'close' to each other. TRPO is an on-policy algorithm. On-policy methods bring many benefits, like the ability to gauge each resulting policy. However, they typically discard all the knowledge about the policies which existed before. In this work, we use a replay buffer to borrow from the off-policy learning setting to TRPO.

Entropy regularization is usually used to improve policy optimization in reinforcement learning. It is thought to aid exploration and generalization by encouraging more random policy choices.

We add an Entropy regularization term to advantage over $\pi$, accumulated over time steps, in TRPO. We call this update "EnTRPO". Our experiments demonstrate EnTRPO achieves better performance for controlling a Cart-Pole system compared with the original TRPO.

## 1. INTRODUCTION

Model-free RL aims to obtain an efficient behavior policy through trial and error interaction with a black-box environment. The goal is to optimize the standard of an agent's behavior policy in terms of the whole expected discounted reward. Model-free RL has a myriad of applications in games [1-3], robotics [4, 5], and marketing [6], to name a few.

Model-free reinforcement learning algorithms do not build a model of the environment. These algorithms are generally applicable, require relatively little tuning, and can easily incorporate powerful function approximates such as deep neural networks.

According to the way how to learn the policy, previous works can be roughly divided into two categories [7]: value-based methods and policy-based methods.

The primary advantage of policy-based approaches, such as REINFORCES is that they directly optimize the quantity of interest while remaining stable under function approximation (given a sufficiently small learning rate). Value-based methods, such as Q-learning, can learn from any trajectory sampled from the same environment. Figure 1 shows the model-free reinforcement learning algorithms classification.

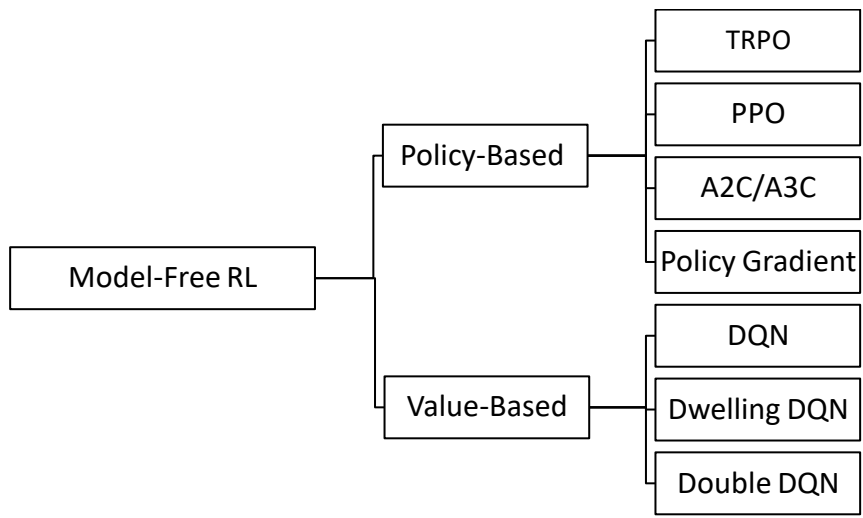

Figure 1- model-free reinforcement learning algorithms classification [8].

In on-policy optimization, vanilla policy gradient algorithms suffer from occasional updates with large step sizes, which cause collecting bad samples. [9]. In order to succeed in such instability, Trust Region Policy Optimization (TRPO) [9] limits the KL divergence between consecutive policies to realize far more stable updates. However, such KL divergence constraint can impose very strict restriction on the new policy iterate, making it difficult to bypass locally optimal solutions and slowing down the learning process.

Recently, some works based on trust regional policies have been proposed. Shani et al. [10] Consider sample-based TRPO and set the $\widetilde{O}(1/\sqrt{N})$ convergence rate to the global optimum. Jha et al. [11] proposed a policy optimization trust region method, which uses Hessian's QuasiNewton approximation, called Quasi-Newton Trust Region Policy Optimization (QNTRPO). Through a series of numerical experiments on challenging continuous control tasks, they demonstrated that their choice is effective in terms of the number of samples and improves performance. Tang and Agrawal [12] proposed the use of normalizing flow policies to improve the trust region policy search. They show that when the trust region is constructed using the KL divergence constraint, the samples generated by the normalized flow policy are far from the "center" of the previous policy iteration, which can achieve better exploration and help avoid bad local optimum. They show that the normalizing flow policy significantly improves the baseline architecture, especially in high-dimensional tasks with complex dynamics. Li and He [13] extended TRPO to multi-agent reinforcement learning (MARL) problems. They show that the policy update of TRPO can be transformed into a distributed consensus optimization problem in a multi-agent scenario. Through a series of approximations to the consensus optimization model, they proposed a decentralized MARL algorithm, which they called multi-agent TRPO (MATRPO). The algorithm can optimize distributed policies based on local observations and private rewards. The agents do not need to know the observations, rewards, policies, or value/action-value functions of other agents. The agents only share the likelihood ratio with their neighbors during the training process. The algorithm is completely decentralized and privacy-preserving. Their experiments in two cooperative games proved the method's strong performance on complex tasks.

In RL, exploration is essential to find good policies in the optimization process: if the optimization process does not sample a large number of different state-action pairs, it may converge to a bad policy. To prevent the policies from becoming deterministic too quickly, researchers use entropy regularization [8]. Its success is sometimes attributed to the fact that it "encourages exploration". Furthermore, the development of well-generalized RL agents is a long-standing challenge that has recently received more attention. There are several known methods to improve generalizability in RL. Choosing an appropriate function approximation model is a way to help generalization between states and actions. Regularization methods can further enhance the generalization. For instance, it is common to regularize the policy space by encouraging policies with high entropy [14]. Our focus is on policy evaluation. To this end, we develop a surrogate objective function by adding entropy regularization. We call this update "EnTRPO". We show that EnTRPO increases reward and lead to an overall improvement due to more stochastic policy changes.

The rest of the paper is structured as follows. The next section reviews background and Section 3 describes the proposed EnTRPO method. Experimental results are reported in Section 4 and the conclusion is presented in Section 5.

## 2. BACKGROUND

We summarize the standard policy gradient framework for RL and we will continue to describe the original TRPO.

The Markov decision process (MDP) comprises of a state space $S$, an action space $A$, a stationary transition dynamics distribution $P: S \times A \times S \to R$, an initial state distribution $\rho_0: S \to R$, a reward function $r: S \times A \to R$, and a discount factor $\gamma \in (0,1)$. In MDP, an agent acts in a stochastic environment $E$ by sequentially choosing actions over a sequence of time steps. At each time step $t$, the agent encounters a state $s_t$ and performs an action $a_t$ according to policy $\pi: S \times A \to [0,1]$. The environment then returns a scalar reward $r(st, at)$ and a new state according to dynamics $P(st+1 \mid st, at)$. According to policy $\pi$, the agent interacts with the MDP to give a trajectory of states, actions, and rewards: $s_0, a_0, r_0, \ldots, s_t, a_t, r_t \ldots$ The return $R_t$ is the cumulative discounted reward from timestep $t$, $R_t = \sum_{k=t}^{\infty} \gamma^{k-t} r(s_k, a_k)$. The state value function and the action value function are defined as the expected return, $v_\pi(st) = E_{a_t, s_{t+1}, \ldots}[Rt]$ and $Q_\pi(s_t, a_t) = E_{s_{t+1}, a_{t+1}, \ldots}$[3] . The formulations of these two functions and their relationship are

$$v_\pi(s_t) = E_{a_t, s_{t+1}, \ldots, \sim \pi} \left[ \sum_{k=t}^{\infty} \gamma^{k-t} r(s_k, a_k) \right] \tag{1}$$

$$Q_\pi(s_t, a_t) = E_{s_{t+1}, a_{t+1}, \ldots, \sim \pi} \left[ \sum_{k=t}^{\infty} \gamma^{k-t} r(s_k, a_k) \right] \tag{2}$$

$$v_\pi(s_t) = E_{a_t \sim \pi(.,s_t)} [Q_\pi(s_t, a_t)] \tag{3}$$

The standard definition for the advantage function then can be represented

$$A_\pi(s_t, a_t) = Q_\pi(s_t, a_t) - v_\pi(s_t) \tag{4}$$

Which provides a relative measure of value with each action since $E_{a_t \sim \pi}[A_\pi(s_t, a_t)] = 0$. Given policy $\pi$, the discounted visitation frequency $\rho_\pi(s)$ can be defined as a discounted weighting of states encountered starting at $s_0$ and then following $\pi$.

$$\rho_\pi(s) = P(s_0 = s) + \gamma P(s_1 = s) + \gamma^2 P(s_2 = s) + \ldots \qquad (5)$$
$$= \sum_{t=0}^{\infty} \gamma^t P(s_t = s | s_0, \pi)$$

In the following, we introduce the TRPO method.

The goal of reinforcement learning is to maximize expected return from the start state, denoted by policy performance objective $\eta(\pi) = E_{s0,a0,\ldots}[R_0]$. This objective is further represented as

$$\eta(\pi) = E_{s_0, a_0, \ldots}[\sum_{t=0}^{\infty} \gamma^t r(s_t, a_{kt})] \qquad (6)$$

Where

$$s_0 \sim \rho_0, a_t \sim \pi(.,s_t), s_{t+1} \sim P(.,s_t,a_t).$$

The expected return of another policy $\tilde{\pi}$ can be expressed in terms of the advantage over $\pi$, accumulated over timesteps

$$\eta(\tilde{\pi}) = \eta(\pi) + E_{s_0,a_0,\ldots \sim \tilde{\pi}}[\sum_{t=0}^{\infty} \gamma^t A_\pi(s_t, a_t)] = \qquad (7)$$
$$= \eta(\pi) + \sum_s \rho_{\tilde{\pi}}(s) \sum_a \tilde{\pi}(a|s) A_\pi(s,a)$$

Note that any policy update $\pi \to \tilde{\pi}$ that satisfies $\sum_s \rho_{\tilde{\pi}}(s) \sum_a \tilde{\pi}(a,s) A_\pi(s,a) \geq 0$ can guarantee the improvement of policy performance $\eta$. The difficulty of such a policy improvement guarantee lies in the dependence of $\rho_{\tilde{\pi}}(s)$ on $\tilde{\pi}$. In order to reduce such dependence, a local approximation to $\eta(\tilde{\pi})$ is proposed by replacing visitation frequency $\rho_{\tilde{\pi}}$ with $\rho_\pi$

$$L_\pi(\tilde{\pi}) = \eta(\pi) + \sum_s \rho_\pi(s) \sum_a \tilde{\pi}(a|s) A_\pi(s,a) \qquad (8)$$

A sufficiently small step $\pi \to \tilde{\pi}$ that improves $L_\pi$ will also Improve $\eta$. However, such an improvement of $\eta$ is limited by this sufficiently small step.

TRPO is proposed to avoid such limitation, which is inspired by the method of conservative policy iteration. TRPO proposes a surrogate objective function based on a lower bound for policy performance _ and proves that maximizing the proposed function can guarantee the improvement of policy performance. Specifically, the lower bound provided in TRPO is as follows:

$$\eta(\pi_{new}) \geq L_{\pi_{old}}(\pi_{new}) - \frac{4\epsilon\gamma}{(1-\gamma)^2} D_{KL}^{max}(\pi_{old}, \pi_{new}) \tag{9}$$

where $\epsilon = \max_{s,a} |A_{\pi_{old}}(s,a)|$, $\pi_{old}$ and $\pi_{new}$ separately represent current policy and new policy, and the distance formulation $D_{kl}^{max}(\pi_{old}, \pi_{new})$ is defined by the Kullback–Leibler (KL) divergence [15]: $D_{kl}^{max}(\pi_{old}, \pi_{new}) = max_s D_{kl}(\pi_{old}(.|s)||\pi_{new}(.|s))$. Equation (9) shows that the improvement of the surrogate objective function in the right-hand side can guarantee the improvement of policy performance η. For this surrogate objective function, its visitation frequency $\rho_{\pi_{old}}$ in $L_{\pi_{old}}(\pi_{new})$ depends on specific behavior policy, i.e., current policy $\pi_{old}$.

## 3. PROPOSED METHOD

As described, the original formulation of the lower bound in TRPO is described in the following

$$\eta(\pi_{new}) \geq L_{\pi_{old}}(\pi_{new}) - \frac{4\epsilon\gamma}{(1-\gamma)^2} \alpha^2 \tag{14}$$

Where

$$\alpha = D_{KL}^{max}(\pi_{old}, \pi_{new})$$

$$\epsilon = \max_{s,a} |A_{\pi_{old}}(s,a)|$$

We add an entropy term $H(\pi(s_t,\cdot))$ to advantage in Equation (7). So the Equation (8) will change to:

$$L'_\pi(\tilde{\pi}) = \eta(\pi) + \sum_s \rho_\pi(s) \sum_a \tilde{\pi}(a|s) A_\pi(s,a) + \alpha\gamma^t H(\pi(s_t,\cdot)) \tag{15}$$

where $\alpha$ is a user-specified temperature parameter that controls the degree of entropy regularization (coefficient for entropy).

The derivation of the surrogate objective function is based on the lower bound in (14). Specifically, we take L' and obtain the following lower bound from (14):

$$\eta(\pi_{new}) \geq \eta(\pi) + \sum_s \rho_\pi(s) \sum_a \tilde{\pi}(a|s) A_\pi(s,a) + \alpha\gamma^t H(\pi(s_t,\cdot)) - \frac{4\epsilon\gamma}{(1-\gamma)^2} \alpha^2 \tag{16}$$

It should be noted that a policy update that improves the right-hand side of (16) can guarantee the improvement of the true performance η. For simplicity, we denote the right-hand side of (16) as $M_\pi(\tilde{\pi})$

$$M_\pi(\tilde{\pi})= \eta(\pi)+\sum_s \rho_\pi(s)\sum_a \tilde{\pi}(a|s) A_\pi(s,a) + \alpha\gamma^t \mathbb{H}(\pi(s_t,\cdot)) - \frac{4\epsilon\gamma}{(1-\gamma)^2}\alpha^2 \quad (17)$$

Note that, when $\tilde{\pi} = \pi$, $M_\pi(\tilde{\pi})$ in (17) can be represented as $M_\pi(\pi)$ in (18). The formula for calculating kl-divergence is $D_{KL}(P||Q) = -\sum_z p(z)\log q(z) + \sum_z p(z)\log p(z) = \mathbb{H}(P,Q) - \mathbb{H}(P)$. Therefore, if $\tilde{\pi} = \pi$, a = 0 and $\mathbb{H}(\pi(s_t,\cdot)) = 0$.

$$M_\pi(\pi)= \eta(\pi)+\sum_s \rho_\pi(s)\sum_a \pi(a|s) A_\pi(s,a) \quad (18)$$

By Equation (4):

$$M_\pi(\pi)= \eta(\pi)+\sum_s \rho_\pi(s)\sum_a \pi(a|s)(Q_\pi(s,a) - V_\pi(s)) \quad (19)$$

By Equation (3):

$$M_\pi(\pi)= \eta(\pi)+\sum_s \rho_\pi(s)\sum_a \pi(a|s)(V_\pi(s) - V_\pi(s)) = \eta(\pi) \quad (20)$$

According to (19) and (20), the lower bound of $\eta(\tilde{\pi}) - \eta(\pi)$ is derived as follows:

$$\eta(\tilde{\pi}) \geq M_\pi(\tilde{\pi}) \quad \text{By Equation (16), (17).} \quad (21)$$
$$\eta(\pi) = M_\pi(\pi) \quad \text{By Equation (20).} \quad (22)$$

Thus

$$\eta(\tilde{\pi}) - \eta(\pi) \geq M_\pi(\tilde{\pi}) - M_\pi(\pi) \quad (23)$$

That is, maximizing the function $M_\pi(\tilde{\pi})$ by choosing appropriate policy $\tilde{\pi}$ can guarantee that the true objective $\eta$ is no decreasing. Therefore $M_\pi(\tilde{\pi})$ is our desired surrogate objective function.

In order to further illustrate the improvement guarantee, we then describe a policy iteration scheme that optimizes the proposed surrogate objective function at each iteration.

$$\pi_{i+1} = argmax_\pi[L_{\pi_i}(\pi) - CD_{KL}^{max}(\pi_i, \pi)] \tag{24}$$

This policy iteration scheme can generate improved policy sequences η($\pi_k$) ≤ η($\pi_{k+1}$) , k∈N.

## 4. EXPERIMENTS

To validate the proposed EnTRPO, we conducted experiments for stabilizing a Cart and Pole. A pole is connected through an un-actuated joint to a cart. the cart moves along a frictionless track. The system is controlled by applying a force of +1 or -1 to the cart. The pendulum starts upright, and the goal is to prevent it from falling over. Every timestep the pole stays upright will receive a reward of +1. The episode ends when the angle between the pole and the vertical exceeds 15 degrees, or the car moves more than 2.4 units from the center.

We first describe the experimental setup that includes the detailed configuration in our method. We next compare the overall performance of our method with the original TRPO method.

*4.1.Setup*

Neural network architecture for policy is the fully connected neural network with two hidden layers of dimension 64 and with Tanh activations. Neural network architecture for state value functions is the fully connected neural network with three hidden layers of dimension 128, 64, 32. The state value network produces a single scalar value to estimate state value. Actor (policy) is updated using GAE[1] method [16] for Advantage estimation. Batch size is 32. Replay memory consists of multiple lists of state, action, next state, reward, and return from state. If the reward is more than 195, the buffer will be cleared.

*4.2.Comparison with TRPO*

For better comparison, the settings (Section 4.1) are the same for implementing both TRPO and EnTRPO methods. Experimentally, we have considered the coefficient for entropy ($\alpha$) equal to 0.0001.

In RL, the convergence is influenced by the discount factor depending on whether it's a continual task or an episodic one. In a continual one, γ must be between [0, 1), whereas an episodic one it can be between [0, 1] [17].

---

[1] Generalized Advantage Estimation

As shown in Figure 2, when discount factor (γ) is equal to 0.8, the convergence of the TRPO and EnTRPO are the same. For γ=0.85, the convergence of both implementation are faster than when γ=0.8. EnTRPO converges before 60 epochs but TRPO converges shortly after 60 epochs. For γ=0.9, TRPO will not converge, but EnTRPO will converge after 120 epochs. It can be seen that EnTRPO has the best performance with a discount factor of 0.85. Therefore, in order to achieve the best performance, we must make a compromise between the discount factor and the coefficient for entropy. This result is the same as result [14]. They have used the $L_2$ regularization in $TD^2$ Learning and have come to the conclusion that they can improve the generalization by properly adjusting the discount factor and the $L_2$ regularization factor.

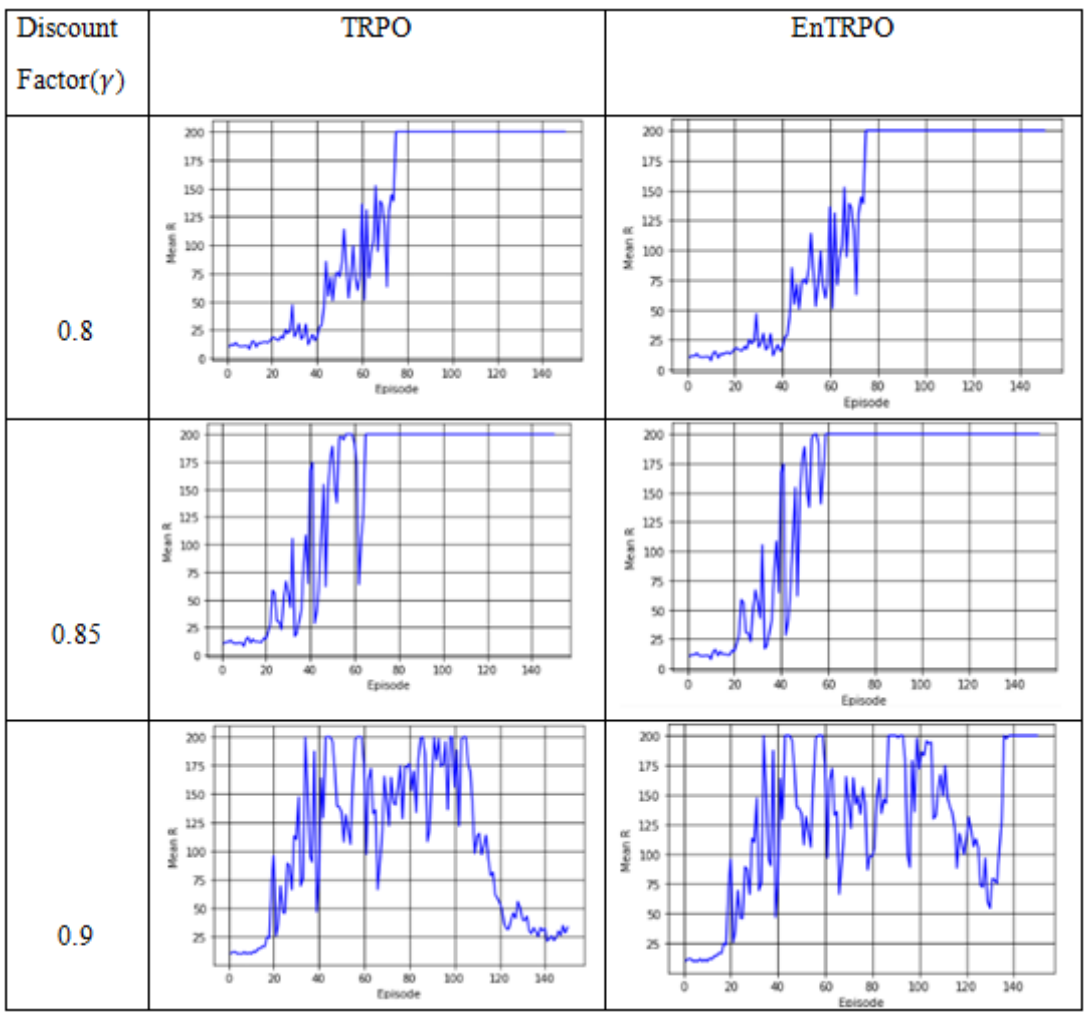

Figure 1- Comparison between EnTRPO and original TRPO.

---

[2] Temporal difference (TD) learning

# 5. CONCLUSION

TRPO updates the policies taking the maximum possible step to improve performance while meeting a particular restriction on how closely new and old policies are allowed. The constraint is expressed in terms of KL-Divergence. This differs from the normal policy gradient, which keeps old and new policies tight in parameter space. TRPO is an on-policy algorithm and can be used for environments with either discrete or continuous action spaces. We develop a surrogate objective function by adding entropy regularization to advantage over π in TRPO. We call this update "EnTRPO". We showed that Convergence speed in EnTRPO is equal or more than TRPO. In addition, when TRPO did not converge, EnTRPO converged. As a future work, it is suggested that the proposed method be tested in more complex environments.